\def\year{2021}\relax
\documentclass[letterpaper]{article} 
\usepackage{aaai21}  
\usepackage{times}  
\usepackage{helvet} 
\usepackage{courier}  
\usepackage[hyphens]{url}  
\usepackage{graphicx} 
\usepackage{natbib}  
\usepackage{caption} 
\usepackage{enumitem}
\usepackage[intlimits]{amsmath}
\usepackage{booktabs}
\usepackage{amsfonts}

\title{Reinforcement Learning-based Product Delivery Frequency Control}
\author {
        Yang Liu,
        Zhengxing Chen, 
        Kittipat Virochsiri, \\
        Juan Wang, 
        Jiahao Wu, 
        Feng Liang \\
}
\affiliations {
    Facebook, Menlo Park, CA, USA, 94025 \\
    \{yliu9, czxttkl, kittipat, juanw, jiahaowu, liangfeng\}@fb.com
}

\begin{document}

\maketitle
\begin{abstract}
  Frequency control is an important problem in modern recommender systems. It dictates the delivery frequency of recommendations to maintain product quality and efficiency. For example, the frequency of delivering promotional notifications impacts daily metrics as well as the infrastructure resource consumption (\textit{e.g.} CPU and memory usage). There remain open questions on what objective we should optimize to represent business values in the long term best, and how we should balance between daily metrics and resource consumption in a dynamically fluctuating environment. We propose a personalized methodology for the frequency control problem, which combines long-term value optimization using reinforcement learning (RL) with a robust volume control technique we termed ``\textit{Effective Factor}''. We demonstrate statistically significant improvement in daily metrics and resource efficiency by our method in several notification applications at a scale of billions of users. To our best knowledge, our study represents the first deep RL application on the frequency control problem at such an industrial scale. 
\end{abstract}

\maketitle

\section{Introduction}

Frequency control is a common and essential problem in recommender systems. A successful recommender system should have careful control of the recommendation delivering frequency in order to foster an engaging interaction with users while consuming infrastructure resources minimally. For example, social media platforms use promotional Emails to remind users of missed information but also aspire to minimize the delivery volume to avoid resource waste or creating spams~\cite{EmailvolumeoptimizatioatLinkedIn,OptimizingEmailVolumeForSitewideEngagement,NotificationVolumeControlandOptimizationSystematPinterest}.  Generally speaking, an increase of  delivery frequency is able to improve product metrics transiently at the cost of infrastructure resources. However, too frequent recommendation may result in user fatigue in the long term \cite{Userfatigueinonlinenewsrecommendation,EmailvolumeoptimizatioatLinkedIn} and potential risks of shutting down of recommendation channels by users~\cite{Notificationsandawareness}. 

Existing works converted the frequency control problem into constrained optimization, putting daily metrics and resource consumption into the objective and constraints separately to counteract with each other~\cite{EmailvolumeoptimizatioatLinkedIn,OptimizingEmailVolumeForSitewideEngagement,NotificationVolumeControlandOptimizationSystematPinterest}. However, the daily metrics were still  defined based on short term (\textit{e.g.}, user activeness in one day) or based on heuristics. In the industry, companies care most about \textit{accumulated} daily metrics and resource consumption over the long term, which we believe should be reflected in the optimization objective in a more principled way. In this light, frequency control should be a sequential decision problem as the user interacts with the recommender system continuously. We illustrate with the following motivating example where previous methods fall short. An inactive user has low interest to recommendations that an constrained optimization algorithm never considers it worth any system resources to deliver recommendation to the user. However, a sufficient amount of recommendations in multiple waves would change the user behavior, turning the user into a regular active one. Clearly, an algorithm that can plan its decisions sequentially will win in the long run. 

This paper proposes a methodology for learning frequency control policies that optimize long-term accumulated measurements with the sequential nature of decisions in mind. We leverage reinforcement learning (RL) to learn the \textit{value} of different frequencies, defined as the best possible accumulated daily metrics and resource consumption since the product is delivered at the frequency. The accumulation measurements represent the potential performance of a system in the long term and are closer to common business metrics companies aim to lift. RL has been demonstrated to be capable of learning long-term values of decisions in complex sequential-decision domains such as games~\cite{minhvideogame} and robotics~\cite{robotics}, where accumulated rewards are more important than the immediate reward at any single step. Take \textit{Go} for an example. It is useful to evaluate a move based on whether it contributes to the final victory but less useful (and sometimes deceptive) to know whether the move brings short-term board advantage. Similarly, the frequency control can also be formulated as a sequential decision problem where a sequence of decisions on delivery frequency would collectively optimize long-term (accumulated) daily metrics and resource usage. In particular, we choose to apply Deep Q-Networks (DQN), an off-policy value-based reinforcement learning algorithm~\cite{Reinforcementlearning} to estimate the values of (user, frequency) pairs. 

Another challenge we aspire to solve is to stabilize global delivery volume at deployment, for which we propose a simple yet robust technique termed ``\textit{Effective Factor}''. In a large-scale industrial setting, we observed that the policy directly derived from the values learned by DQN could result in noticeable fluctuation in global delivery volume because the user and system behavior could shift day by day in the real world. While the data shift is not drastic generally, the change in delivery volume can be significant enough to trigger alerts from other monitoring systems present in modern platforms. Even recurrent training cannot help RL models catch up with the latest data shift because RL models train on user history data spanning a few weeks. Therefore, we propose Effective Factor which monitors global delivery volume and acts on top of DQN to stabilize output frequencies dynamically.

\textbf{Contributions}. Our study represents:
\begin{itemize}
\item Formulate the frequency control problem as a sequential decision problem.
\item Propose to solve the frequency control problem by reinforcement learning and Effective Factor (dynamic volume control).
\item Evaluate our method and show positive gains in both daily metrics and resource savings in several applications at a scale of billions of users.
\end{itemize}

To our best knowledge, our work has been the first deep RL-based algorithm for the frequency control problem that has been validated by experiments at a scale of billions of users. The rest of the paper is outlined as follows. We first introduce related works. Then, we propose our methodology in learning long-term values and stabilizing frequency control policies using ``\textit{Effective Factor}'' (\textit{Methodology} Section). We then evaluate our method in several real-world recommendation applications (\textit{Applications and Experiments} Section). Importantly, the proposed methodologies can be easily applied to other frequency control cases with minor modifications.

\section{Related Work}
\label{related_work}

The works most relevant to ours come from large-scale social media platforms (\textit{e.g.}, LinkedIn and Pinterest), which involved constrained optimization balancing between delivery volume and daily metrics~\cite{EmailvolumeoptimizatioatLinkedIn,OptimizingEmailVolumeForSitewideEngagement,NotificationVolumeControlandOptimizationSystematPinterest}. Gupta~\textit{et al.} frames the frequency control problem as a Multi-Objective Optimization (MOO) problem~\cite{EmailvolumeoptimizatioatLinkedIn}. The objective is to minimize the expected delivery volume as the sum of individual emails' delivery probabilities. The objective comes with the constraints mandating that positive/negative user experience should be above/below certain thresholds. Later, they extended the constraints to include sitewide engagement, such as the total number of active users on a platform~\cite{OptimizingEmailVolumeForSitewideEngagement}. The solution can be obtained from a large-scale quadratic programming solver. However, their formulation indicates a strong independence assumption between emails, which is unable to represent the sequential effect of recommendation delivery in real-world applications. Zhao~\textit{et. al} proposed another constrained optimization technique which relaxes the independence assumption between emails~\cite{NotificationVolumeControlandOptimizationSystematPinterest}. Under a total volume constraint, they searched for the best frequency $f$ for each user $u$ that contributes most $p(a|u,f)$, the conditional probability of $u$ being active. $p(a|u,f)$ is directly predicted by supervised learning models based on user features and the delivery frequency. However, it remains as heuristics as for what constitutes the user activeness label. More importantly, the delivery decisions are still made independently, without considering that a sequence of decisions could change daily metrics differently .

Reinforcement learning (RL) has been successful in learning policies for sequential decision problems in various domains such as video games~\cite{minhvideogame,starcraft}, board games~\cite{go}, real-time bidding~\cite{DQNbasedRTB}, and robotics~\cite{robotics}. Its strength of capturing long-term effect also attracts researchers working on recommender systems to use RL for retrieving engaging items~\cite{pagewise_rec, drn_news, ads}. In the notification domain, the reinforcement learning approach has achieved success recently~\cite{Reinforcementlearningapplications}, in a similar motivation of viewing recommendation as a sequential decision problem. Gauci~\textit{et al.} reported successful applications of an RL-based policy in the notification sending framework to improve daily metrics~\cite{Horizon}. However, the application exclusively focused on making send/drop decisions for individual notifications but did not consider how a learned policy would interact with infrastructure resources such as delivery volume. 

There is a relevant branch of research on frequency capping for advertisements~\cite{Onlinestochasticmatching,Privacypreservingfrequencycappingininternetbanneradvertising,DeterminingoptimaladvertisementfrequencycappingpolicyviaMarkovdecisionprocessestomaximizeclickthroughrates}. The goal of frequency capping for an ads server is to provide a supply-demand matching between advertisers who want to avoid repeated displays to the same user and users who arrive sequentially in an unknown pattern. However these studies share a different goal than us in that they put caps on frequencies per user rather than on the global delivery volume of a system.

\section{Methodology}
\label{methodology}

\subsection{Problem Definition}

As we stated in \textit{Introduction}, the core idea of frequency control is to keep a right pace of recommendation delivery in order to maintain accumulated daily metrics as well as minimize resource consumption. While increasing delivery frequency can usually boost short-term product metrics, it is at the cost of infrastructure resource consumption and long-term user experience. Therefore, frequency control should be considered as a sequential decision problem where product delivery at a series of frequencies determines overall user experience and resource consumption together. We formalize the frequency control problem as a Markov Decision Process (MDP)~\cite{AMarkoviandecisionprocess}, a classic framework for solving sequential decision problems. Specifically, the MDP is designed with the following ingredients:
\begin{itemize}
    \item state space $\mathcal{S}$, which defines user features such as the user's profile and historical interactions with the product. 
    \item action space $\mathcal{F}$, which defines the possible frequencies to deliver recommendation. In this paper, we focus on a finite action space with a set of predefined frequencies. Recommendation will be delivered at one of these predefined frequencies each time a decision is due.
    \item reward function $R: \mathcal{S} \times \mathcal{F} \rightarrow \mathbb{R}$, which denotes the quantitative measurement of daily metrics and resource consumption obtained by delivering recommendation $f \in \mathcal{F}$ times to the user with $s \in \mathcal{S}$. In our work, we use a linear reward function: $R(s, f) = m_u(s) - \epsilon f$. It encourages daily metrics $m_u(s)$ while penalizing linearly with the delivery frequency because we want to keep down resource consumption.
    
    \item transition function $T: \mathcal{S} \times \mathcal{F} \rightarrow \mathcal{S}$, which reflects how the user state would adapt after recommendation is delivered at frequency $f$ to the user.
\end{itemize}

The optimal frequency control policy maximizes the accumulated discounted rewards since each user state $s_t$:

\begin{equation}
\label{optimal_policy}
\pi^* = argmax_\pi \; \mathbb{E}_{s_i, f_i \sim \pi}\left[ \sum\limits_{i=t}^\infty\gamma^{i-t} R(s_i, f_i)\right],
\end{equation}

where $\gamma$ is a discount factor which balances the focus of the policy between near-term and longer-term rewards.

\subsection{Deep Q-Network for Value Learning}
Q-learning is a classic RL algorithm in which it seeks for $\pi^*$ by first learning the state-action value, $Q^*(s_t,f_t)$:
\begin{equation}
    Q^*(s_t,f_t)=max_\pi Q^\pi(s_t,f_t)
\end{equation}

\begin{equation}
    Q^\pi(s_t,f_t) = R(s_t, f_t) + \mathbb{E}_{s_i, f_i \sim \pi}\left[\sum\limits_{i=t+1}^\infty\gamma^{i-t} R(s_i, f_i) \right] 
\end{equation}

$Q^*(s_t, f_t)$ is the maximal possible accumulated rewards that could be obtained by any policy. Intuitively, $Q^*(s_t, f_t)$ measures how promising delivering recommendation with $f_t$ frequency to a user $s_t$ would lead to the best long-term gain. Therefore, we treat $Q^*(s_t, f_t)$  as a good proxy of the long-term value of $f_t$ for the user state $s_t$. 

One can use the learned $Q^*(s_t, f_t$) values to derive the optimal policy equivalent to Eqn.~\ref{optimal_policy}:

\begin{equation}
\label{q_optimal_policy}
 \pi^*(s_t) = argmax_{f_t} Q^*(s_t, f_t) 
\end{equation}

Q-learning uses the Bellman update to learn the $Q^*$ function (with $\alpha$ as the learning rate):
\begin{equation}
\begin{split}
    \hat{Q}(s_t,f_t)= (1 - \alpha) \hat{Q}(s_t,f_t) + \\ \alpha \left(R(s_t,f_t) + \max_{f_{t+1}} \hat{Q}(s_{t+1}, f_{t+1}) \right)
\end{split}
\end{equation}

Since our state space contains real-world user features which can be high-dimensional, we choose to train with Deep Q-Network (DQN) \cite{Human-levelcontrolthroughdeepreinforcementlearning} with deep neural networks as function approximation of $\hat{Q}(\cdot,\cdot)$. We use standard training techniques to train the Double Q-Learning~\cite{double_network} such as Dueling Networks~\cite{dueling_network} in order to make learning easier and more stable~\cite{Horizon}.

\begin{figure}
\centering
\begin{minipage}{.8\linewidth}
  \includegraphics[width=\linewidth]{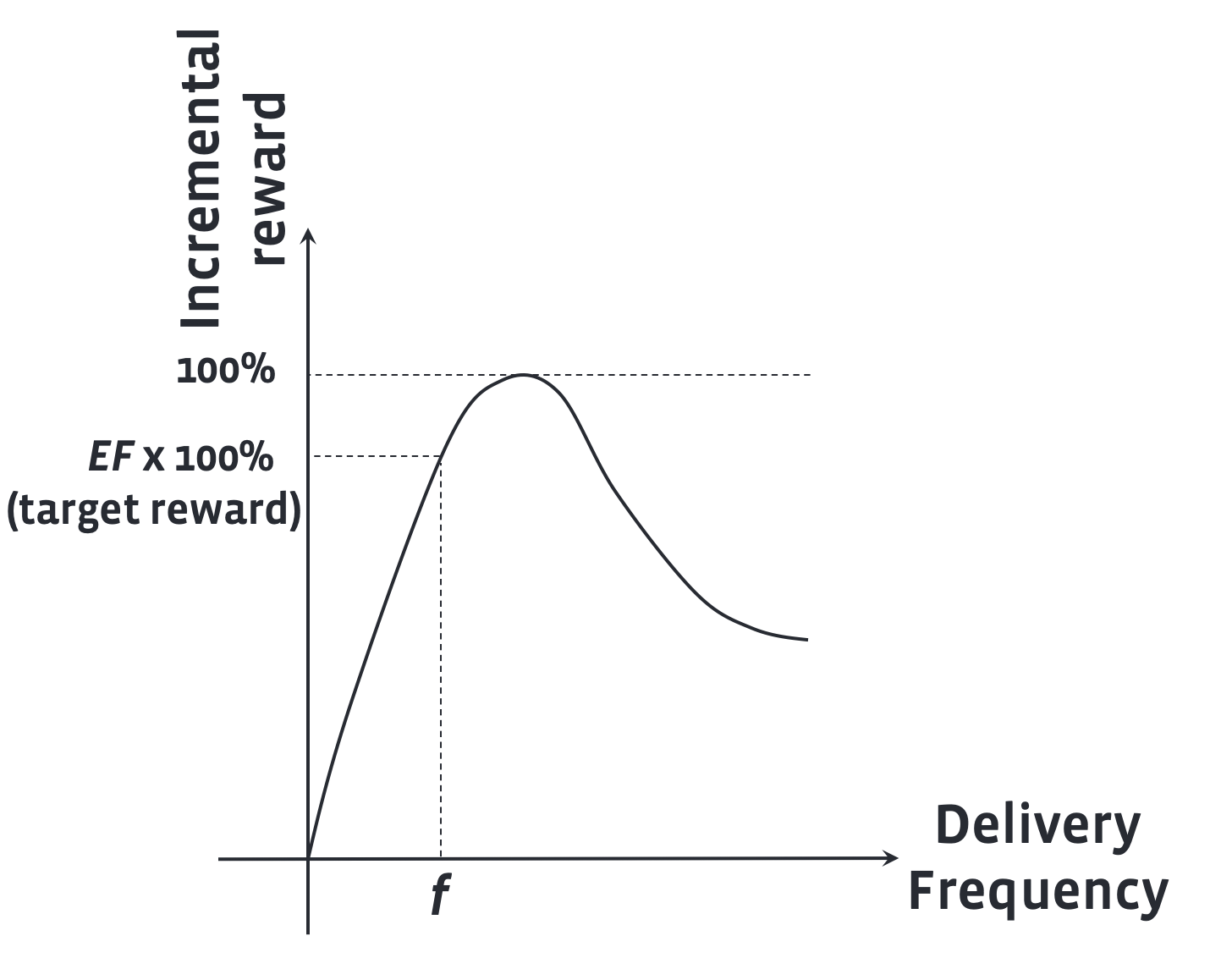}
  \captionof{figure}{Illustration of the \textit{Effect Factor} technique. The incremental reward (y-axis) by the selection of various delivery frequency (x-axis) is quantified by $\left[ Q^*(s,f) - min_{f'} Q^*(s,f')\right] / \Delta Q^*(s) $. The returned delivery frequency for each user in the online serving is the minimal frequency that achieves $min_{f'} Q^*(s,f') + EF\cdot\Delta Q^*(s)$. }
  \label{fig:ef_lines}
\end{minipage}
\hspace{.02\linewidth}
\newline
\newline
\hfill \break
\hfill \break
\begin{minipage}{.8\linewidth}
  \includegraphics[width=\linewidth]{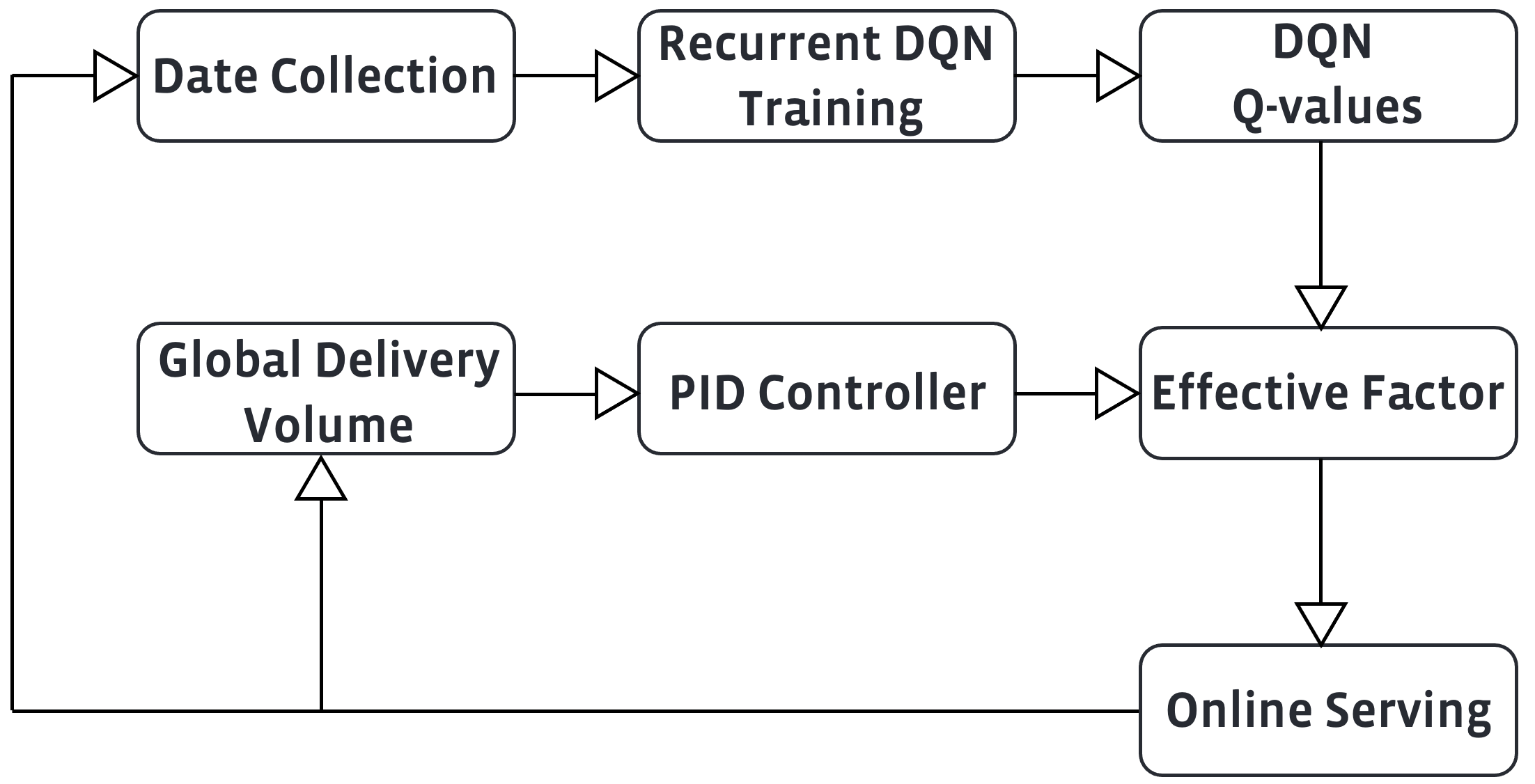}
  \captionof{figure}{System Overview. The system consists of (1) a data pipeline to collect RL training data; (2) a recurring training pipeline of DQN models on a daily basis; (3) an effect factor controller, which reads predefined configurations, monitors real-time global volume, and adjusts dynamically; and (4) an online serving system. In the RL application, the clients provide either a fixed value of effect factor or a predefined target delivery volume. If latter, a PID controller will be applied to adjust the effect factor in online serving automatically.}
  \label{fig:overview}
\end{minipage}
\end{figure}

\subsection{Effective Factor}
Although in theory we can use a trained DQN and Eqn.~\ref{q_optimal_policy} as a serving policy, it often results to jittering global delivery volume as real-world environments are shifting dynamically. The shifts in user behaviors or system resource usage are generally slow but still significant. We hypothesize the shifts come from temporal effects (\textit{e.g.}, more traffic on the weekend or holidays than a weekday) and internal changes (\textit{e.g.}, changes in the upstream/downstream services, throttles/constraints newly enforced into the system). In our real-world preliminary tests we found that the global delivery volume could fluctuate by a low but significant level ($10\sim 20\%$) to trigger alerts from in-house monitoring systems.

While companies would like to see long-term gain in daily metrics and resource consumption, they also hope to stabilize global delivery volume. Recurrent training cannot help RL models catch up the latest user/system behavior shift because RL models train on user history data spanning a few weeks. It is also not easy to massage the reward function, particularly $\epsilon$ which we use to control the impact of resource consumption, to get a stable global delivery volume. Because the relationship between recommendation frequency and global resource consumption may not be easily defined as a linear function.  

We propose a simple yet robust technique called ``\textit{Effect Factor}'' on top of Q-values to stabilize global delivery volume in real time. The technique works for our cases where the action space has at least 3 predefined frequencies\footnote{When the action space has only two possible actions, the practitioner can design a simpler heuristic.}. First, We define the maximal incremental value of Q-values of a state as:
\begin{equation}
\Delta Q^*(s) = max_{f'} Q^*(s,f') - min_{f'} Q^*(s,f')
\end{equation}
Then, we define an adjustable scalar $EF \in [0, 1]$ and a frequency control policy based on $EF$ and $\Delta Q^*(s)$:

\begin{multline}
\pi^{EF}(s) = \\ \min\left\{f \;|\; Q^*(s,f) \geq min_{f'} Q^*(s,f') + EF \cdot \Delta Q^*(s)\right\},
\end{multline}
where $\pi^{EF}(s)$ can be intuitively explained as finding the minimal frequency that achieves a sufficient level of the maximal incremental value.

We observed in practice that tuning $EF$ has a predictable effect on daily metrics and resource consumption: for most states, as $f$ increases, $Q^*(s,f)$ would be monotonically increasing, monotonically decreasing, or vary with a bell shape (See Fig.~\ref{fig:ef_lines} for example). Although adjusting $EF$ could result to slow change for daily metrics and resource consumption of a single user, from a global view adjusting $EF$ brings very immediate and monotonic change in terms of the overall delivery volume.

Since $EF$ is a responsive and monotonic knob for tuning the global delivery volume, we employed a proportional-integral-derivative (PID) controller \cite{PIDcontrollers} to dynamically adjust $EF$ in order to stabilize the global delivery volume. To guide directions and magnitudes of adjustment, PID controllers calculate the proportional, integral, and derivative of the differences between a target delivery volume and actual delivery volumes. To run the PID controllers, a client team will first set a global target delivery volume in a configuration file. We have a service running continuously to aggregate and store the actual global delivery volume. A recurring job is periodically (\textit{e.g.}, per 10-min) executed to check the difference between the actual and target delivery volume, based on which $EF$ is dynamically adjusted. We observed that PID controllers worked as expected: whenever the actual global delivery volume exceeds the expected volume, PID controllers will lower the $EF$; if the actual volume goes in the opposite direction, PID controllers will increase the $EF$ instead. We can also assign different values of $EF$ for different user cohorts based on product needs.

It is worth noting that although the $EF$-intervened policy would deviate from the optimal policy specified in Eq.~\ref{q_optimal_policy}, $EF$ is set in practice to cause only very mild deviation, and we can still see gains in metrics we would like to lift.

\section{Applications and Experiments}
\label{experiments}

We built a personalized frequency control system to support our RL-based methodology described in the \textit{Methodology} Section. An overview of our system can be seen in Figure~\ref{fig:overview}. We use the ``notification scheduling'' domain as a test bed for our RL-based methodology. We believe our method can be extended to other applications that need delivery frequency control in general. Notifications is an important recommendation channel in social network products to deliver missed information to users. While some notifications are triggered by user activities (\textit{e.g.}, direct messages), we focus on those scheduled daily to target users for promotional purposes (\textit{e.g.}, reminding users of missed content or suggesting new friends to connect). The notifications are scheduled in advance (\textit{e.g.} one day ahead), at which time we need to determine how many notifications to schedule on the delivery day. Our scheduling frequency ranking platform is only responsible for determining the frequency of notification scheduling. The actual content of scheduled notifications is determined by downstream rankers, which can be seen as part of the MDP environment of which the frequency control policy has no direct control.  

\subsection{Training}
To have a comprehensive evaluation of our methodology, we tested on five different notification types which serve different products and run in different channels such as email and mobile push. For business reasons, we will not reveal exact notification types but denote them as $notif\_type_1, \ldots, notif\_type_5$, each of which reaches billion-scale users. We allocated a 0.5\% user segment traffic for training data collection. Whenever a scheduling decision was due for a user, one of six predefined frequencies was chosen to apply (\textit{i.e.}, $\mathcal{F} = \{f_1, f_2, \ldots, f_6\}$). We joined data over user identities such that each user's complete scheduling history (30 tuples of $(s_t, f_t, R(s_t, f_t), s_{t+1})$ ) are included in the final training data. There are totally 50 ranking features used for the model training and prediction, including user-level features and user-notification interaction features. We used an open-source applied reinforcement learning platform \textit{ReAgent} to train the DQN model~\cite{Horizon}.

The hyperparameters (\textit{e.g.}, neural network size, learning rate, \textit{etc.}) of the DQN model are hand-picked without too much fine-tuning. The discount factor $\gamma$ is tuned in the range from 0.25 to 0.9 based on the online experiment results. The daily metric measurement in the reward, $m_u(s)$, is itself a linear function of two different daily metrics. The linear weights of the daily metrics and $\epsilon$ appeared in the reward function, as well as the target global delivery volume, were hand-tuned after 2 to 3 experiment iterations of online A/B tests to achieve accepted trade-offs between daily metrics and resource consumption. In the first iteration, we began with $\epsilon = 0.005 \times daily\ metrics$ and searched its neighborhood. In the next iterations, we searched with greater granularity around the $\epsilon$ that gave us the most acceptable trade-off from the previous iterations. The agreement of trade-offs was reached by discussing with other stakeholders such as product managers.

\subsection{Results}
\begin{figure}
\centering
\begin{minipage}{.8\linewidth}
  \includegraphics[width=\linewidth]{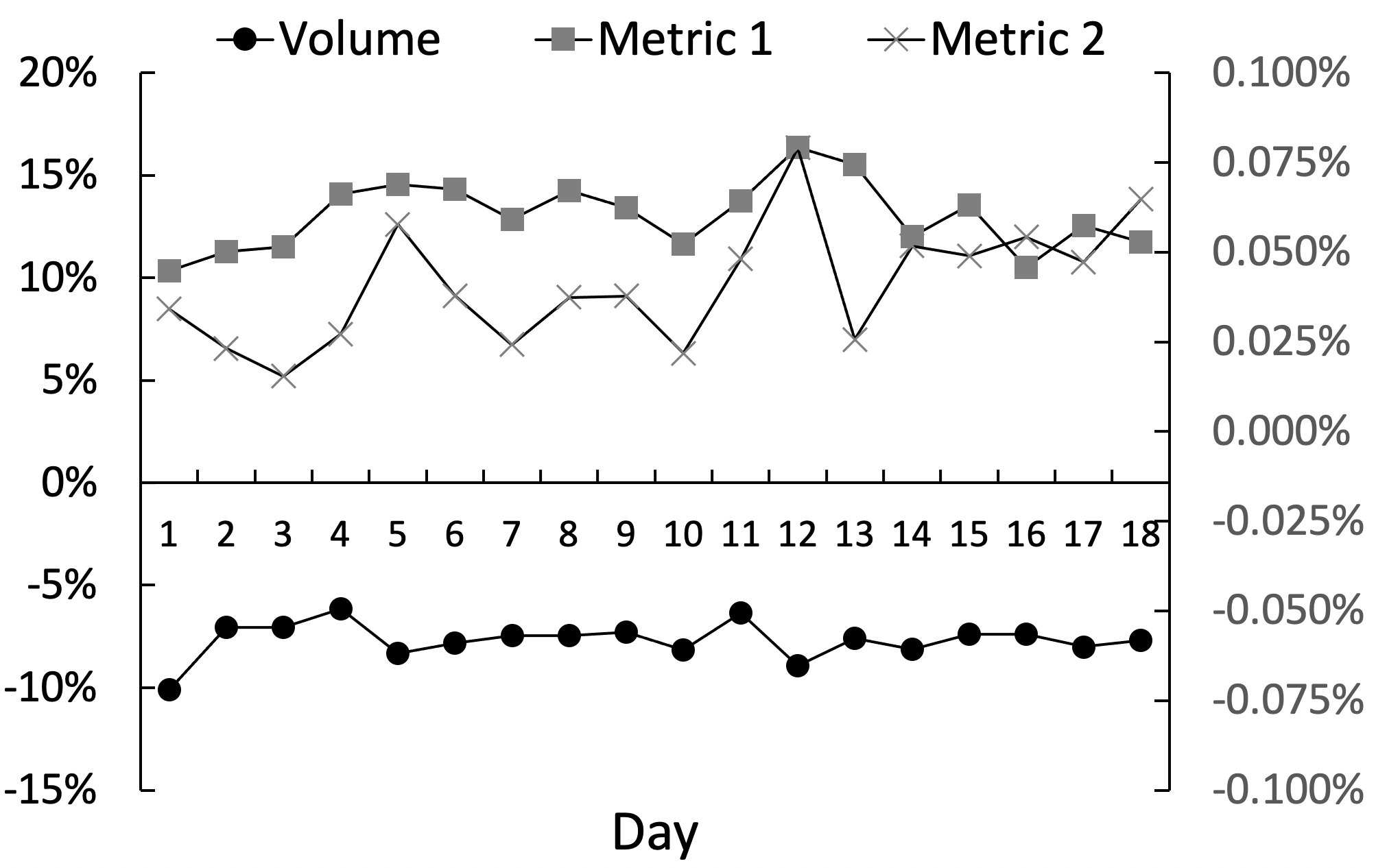}
  \captionof{figure} {Time series of schedule volume and two daily metrics. The latter two are linearly combined in $m_u(s)$ in the reward function.}
  \label{fig:time_series_3_reward}
\end{minipage}
\hspace{.01\linewidth}
\newline
\newline
\hfill \break
\hfill \break
\begin{minipage}{.8\linewidth}
  \includegraphics[width=\linewidth]{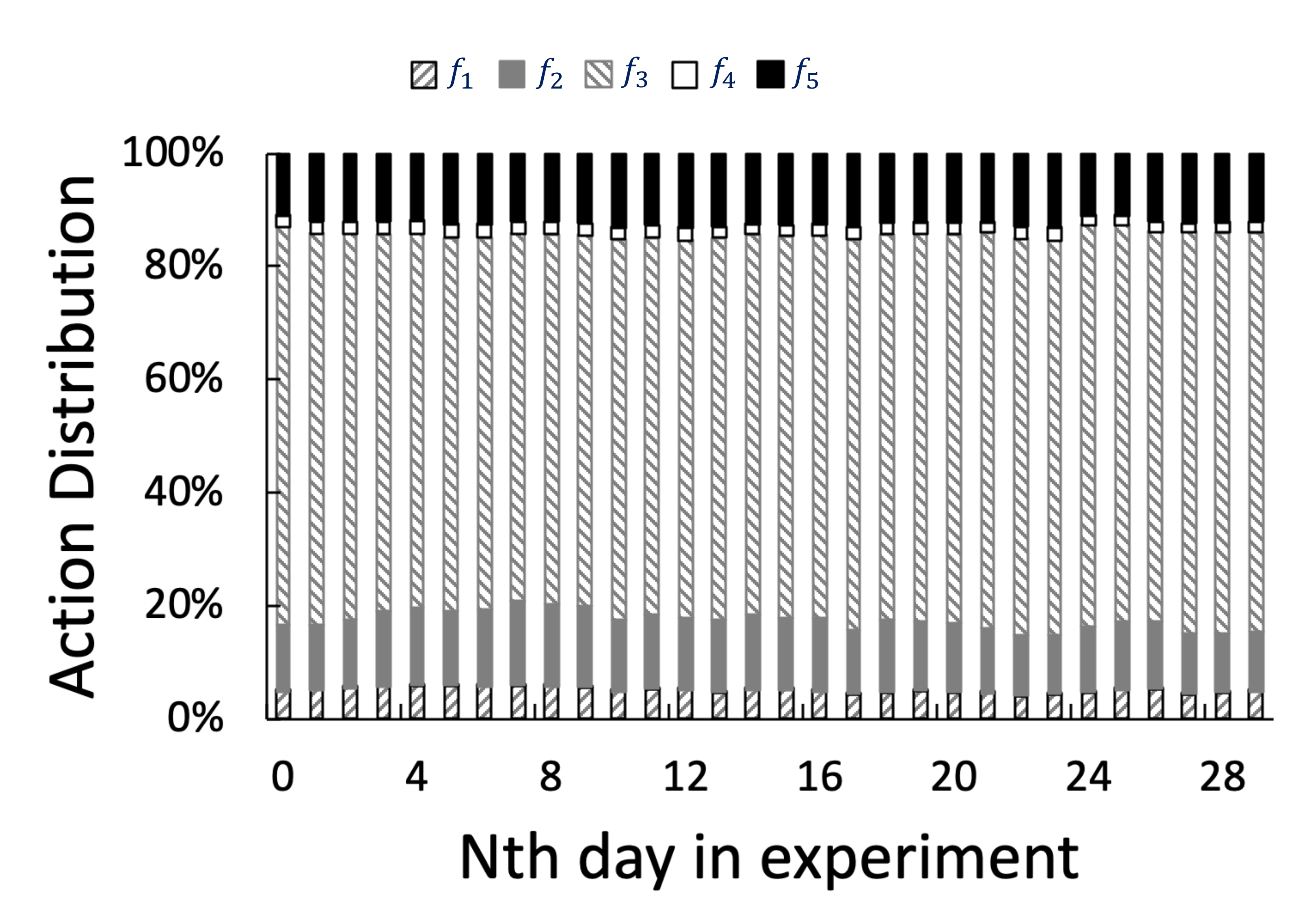}
  \captionof{figure}{Global frequency distribution over the experiment period.}
  \label{fig:freq_dist}
\end{minipage}
\hspace{.01\linewidth}
\newline
\newline
\hfill \break
\hfill \break
\begin{minipage}{.8\linewidth}
  \includegraphics[width=\linewidth]{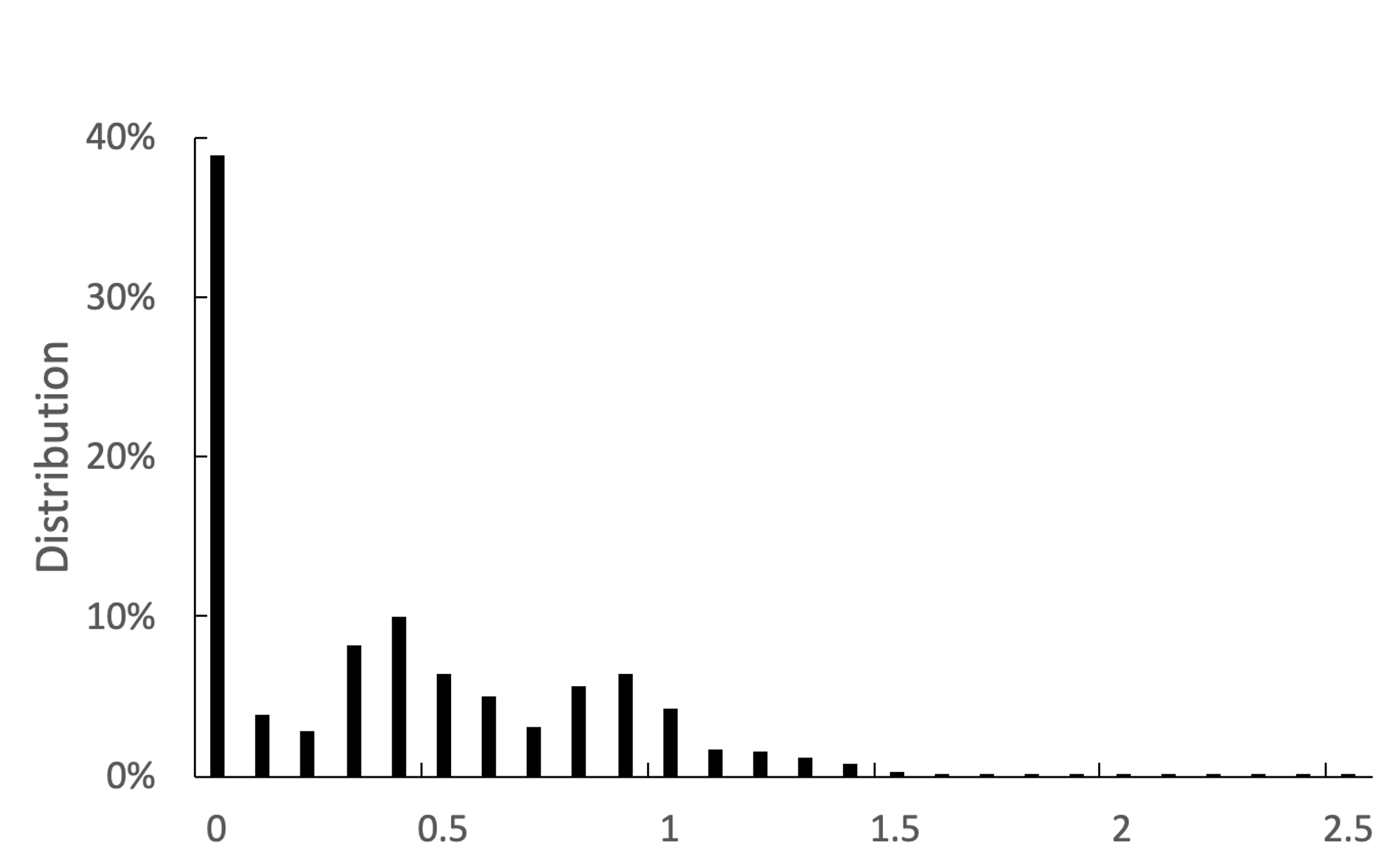}
  \captionof{figure}{Standard deviation of frequencies per user over the 30 days. Around 40\% users received a constant frequency while the rest received varied frequencies under our policy.}
  \label{fig:std_per_user_freq}
\end{minipage}
\end{figure}

The online experiments in this study were performed on the 2-4\% (around 50 million users for each notification type) of the whole monthly active users ($\sim$ 2.5 billion). We compared the performance between our RL-based approach and the current rule-based frequency control product policy. We observed that the RL-based approach has a significantly better efficiency ratio (=daily metrics / delivery volume) than a human-crafted rule-based approach across all the five notification types (Table~\ref{result-table}). In addition, we tracked the two daily metrics included in $m_u(s)$ and the delivery volume, which showed consistent trends over time (See Figure~\ref{fig:time_series_3_reward} for such analysis from one notification type). This shows the possibility that we can improve individual metrics collectively by a linear reward shaping. In addition, we observed that reward functions without $\epsilon$ penalty results in lower delivery efficiency ($daily\ metrics\ /\ delivery\ volume$) and we have to end the experiment prematurely to avoid further negative impact. 

\begin{table*}[h!]
	\centering
	\begin{tabular}{c|ccccc}
		\toprule
		 & $notif\_type_1$ & $notif\_type_2$ & $notif\_type_3$ & $notif\_type_4$ & $notif\_type_5$\\
		\midrule
		delivery volume &  -5.0\% & -7.6\% & +0.9\% &  -5\%  & +6.3\%\\
		daily metrics & + 6.6\% & + 11.7\%  & +17.2\% & Neutral & +12.3\%\\
		\midrule
		efficiency ratio & +12\% & +19\% & +16\% & +5\% & +6\% \\
		\bottomrule
	\end{tabular}
	\caption{Performance of our methodology compared to the current production setting. It achieves higher efficiency ratio\linebreak  (=daily~metrics~/~delivery~volume) in all notification types.}
	\label{result-table}
\end{table*}

\begin{table}[h!]
	\centering
	\begin{tabular}{c|ccc}
		\midrule
		 Activity cohort & High & Medium & Low \\
		\midrule
		Delivery volume &  -20.6\% & -15.9\% & -10.8\%\\
		\midrule
		Daily Metric 1 &  +12.3\% & +16.5\% & +3.8\%\\
		\midrule
		Daily Metric 2 &  +0.0419\% & +0.0297\% & +0.1089\%\\
		\bottomrule
	\end{tabular}
	\caption{User cohort analysis for one notification type. Results are compared to the current production policy.}
	\label{cohort-table}
\end{table}

To have a better understanding of our methodology, we made a case study on one chosen notification type with data tracked over 30 days. The global delivery volume during the experiment period is stable (not shown). Figure~\ref{fig:freq_dist} shows the frequency distribution, which exhibited small fluctuation for each frequency. At the user level, we found that around 40\% users always received the same predicted frequency number and the other users received fluctuated predicted frequency over time, as indicated by the standard deviation of received frequencies per user (Figure~\ref{fig:std_per_user_freq}). Breaking down users into three cohorts (highly, medium, and lowly active) according to their profiles \textit{during} the experiment, we found reduction in the sending volume and improvement in individual daily metrics (the two used in the reward function) in all user cohorts (Table~\ref{cohort-table}).

We also investigated the characteristics of $EF$. We found that once client teams set a reasonable target global delivery volume, $EF$ were dynamically adjusted by PID controllers mostly in the range from 0.75 to 1.0, which indicates that the $EF$-modified policy is still close to the optimal policy learned by DQN. 

\section{Conclusion}
\label{others}
In this paper, we proposed a methodology for the frequency control problem, combining long-term value learning by deep reinforcement learning and a global delivery volume control technique termed \textit{Effective Factor}. The experiment results demonstrate that the proposed approach is able to improve daily metrics effectively while reducing infrastructure resource consumption in the notifications domain at the industrial scale. To our best knowledge, our work represents the first industrial application of deep reinforcement learning in the frequency control problem at a scale of billions of users. 

\section{Acknowledgments}
We thank Yuankai Ge and Jason Gauci's support in this project. We thank all partners and client teams' collaboration and contributions.

\medskip
\bibliography{ref}
\end{document}